\documentclass{article}

\usepackage{graphicx}
\usepackage{amsmath, amssymb}
\usepackage{authblk}
\usepackage{float}
\usepackage{adjustbox}
\usepackage{xcolor}         
\usepackage{url}
\usepackage[preprint]{neurips_2025}

\newcommand{\necessity}[1]{\Box #1}
\newcommand{\possibility}[1]{\Diamond #1}

\newcommand\blfootnote[1]{%
  \begingroup
  \renewcommand\thefootnote{}\footnote{#1}%
  \addtocounter{footnote}{-1}%
  \endgroup
}

\title{Agentic System with Modal Logic for Autonomous Diagnostics}

\author{Antonin Sulc\thanks{asulc@lbl.gov}~}
\author{Thorsten Hellert}
\affil{Lawrence Berkeley National Lab, U.S.A.}

\begin{document}
\blfootnote{Code available at https://github.com/sulcantonin/neuro-symbolic-diagnostics.git}
\maketitle

\begin{abstract}
The development of intelligent agents, particularly those powered by language models (LMs), has shown a critical role in various environments that require intelligent and autonomous decision-making. 
Environments are not passive testing grounds, and they represent the data required for agents to learn and exhibit in very challenging conditions that require adaptive, complex, and autonomous capacity to make decisions. 
While the paradigm of scaling models and datasets has led to remarkable emergent capabilities, we argue that scaling the structure, fidelity, and logical consistency of agent reasoning within these environments is a crucial, yet underexplored, dimension of AI research. 
This paper introduces a neuro-symbolic multi-agent architecture where the belief states of individual agents are formally represented as Kripke models. This foundational choice enables them to reason about known concepts of \emph{possibility} and \emph{necessity} using the formal language of modal logic. 
In this work, we use immutable, domain-specific knowledge to make an informed root cause diagnosis, which is encoded as logical constraints essential for proper, reliable, and explainable diagnosis. 
In the proposed model, we show constraints that actively guide the hypothesis generation of LMs, effectively preventing them from reaching physically or logically untenable conclusions. In a high-fidelity simulated particle accelerator environment, our system successfully diagnoses complex, cascading failures by combining the powerful semantic intuition of LMs with the rigorous, verifiable validation of modal logic and a factual world model and showcasing a viable path toward more robust, reliable, and verifiable autonomous agents.
\end{abstract}

\section{Introduction}
 
The trajectory of progress within the field of artificial intelligence has been influenced by the principles of scaling laws. 
The practice of scaling model size, dataset volume, and available computation has successfully allowed a wide range of emergent capabilities in large language models~\cite{wei2022emergent}. 
In a parallel intellectual movement, the field of agentic AI shows that a similar scaling paradigm must be rigorously applied to the environments in which agents operate~\cite{sapkota2025ai}. The availability of rich, interactive, and diverse environments is foundational for the development of agents that can reason and act with true autonomy. 
As the research focus shifts from agents trained on static datasets to those learning through sustained, dynamic environmental interaction via reinforcement learning (RL), the environment itself is transformed into the primary source for agent learning.

Simply making an environment look more realistic or adding more complexity is not enough to create the next generation of truly capable AI agents. For these agents to operate independently and reliably, especially in high-risk areas like industrial process control, automated scientific research, and autonomous accelerator control~\cite{sulc2024towards}, we need to understand how they think, be able to formally check their reasoning, and ensure their decision-making is firmly based on the core principles of the field they work in.  Language Models (LMs) are powerful tools for processing and generating language, but they sometimes produce answers that sound convincing yet are factually wrong or logically flawed. This problem, often called hallucination, makes it risky to use them in critical systems where safety and reliability are essential \cite{ji2023survey, zhang2025siren}.

We propose a framework that addresses this pressing challenge by integrating two powerful computational paradigms: the creative hypothesis-generation capabilities of LMs and the strict, formal verification of symbolic logic. Specifically, we introduce a multi-agent architecture where each agent's internal state is not a simple state vector but a formally structured Kripke model \cite{kripke1963semantical}. This architectural choice allows agents to maintain and reason about a set of possible worlds, using the operators of modal logic $\necessity{}$ for necessity and $\possibility{}$ for possibility to formalize their beliefs about the environment's current state and its potential future evolutions.

The core contributions of this work establish a comprehensive approach to building more reliable agents in high-stakes environments. First, we present a complete neuro-symbolic agent architecture where LMs are strategically employed for semantic interpretation and hypothesis generation, while modal logic provides the formal backbone for belief representation and state transition.

Second, we introduce a powerful mechanism for injecting expert knowledge into the system as a set of logical axioms. These axioms act as strict constraints to prune the hypothesis space of the LM and ensure its reasoning aligns with established physical laws or operational doctrines.

Third, we have developed a hierarchical diagnostic process that combines (1) causal hypothesis generation, (2) rigorous logical validation against these expert rules, and (3) factual verification against a knowledge base that represents the environment's physical ground truth.

We demonstrate the efficacy of this architecture within a simplified simulation of a particle accelerator control system. 
The domain is characterized by complex, cascading failures where correct and timely root cause analysis is absolutely critical. 
Our simulation results show that this neuro-symbolic approach correctly and reliably distinguishes between causal and merely correlational events—a sophisticated feat of reasoning that remains exceptionally challenging for purely neural approaches. While these findings are a significant step forward in the research direction outlined in~\cite{sulc2024towards}, a real-world proof of concept will be necessary for full validation.

\section{Theoretical Framework: Modal Logic and Neuro-Symbolic Beliefs}
 
To construct agents that can reason robustly and reliably under conditions of uncertainty, we must first equip them with a formal language that can express concepts extending beyond simple propositional truth. An agent operating within a dynamic environment must be capable of reasoning not only about what is currently true, but also about what must be true, what might possibly be true, and, crucially, what cannot be true. Modal logic provides the formal, well-understood mathematical tools required for precisely this kind of sophisticated reasoning.

\subsection{Kripke Models as Agent Belief States}

At the very heart of our proposed architecture, each agent's internal belief state is represented by a Kripke model, formally defined as a tuple $M = (W, R, V)$. This structure is composed of three key elements:

 First, $W$ stands for a set of possible worlds. For a diagnostic agent, a "world" represents a complete and plausible state of the entire environment. For instance, a world $w_0$ might represent a state where the system is operating nominally, while another world $w_1$ could represent a state where a specific cooling fault has occurred.
The second element is $R$, which denotes an accessibility relation defined on $W$. A relation $(w_i, w_j) \in R$ signifies that if the agent currently believes it is in world $w_i$, it considers world $w_j$ to be a possible future or alternative state, thereby defining the agent's uncertainty and its internal model of state transitions.
Lastly, $V$ is a valuation function that assigns a set of true atomic propositions to each world $w \in W$. For example, the valuation of world $w_1$ might be $V(w_1) = \{\text{pressure\_low}, \text{cooling\_fault\_reported}\}$.

An agent's current belief is always anchored to a specific world, $w_{\text{current}} \in W$. The agent then reasons about its beliefs by evaluating modal operators against this model. 
A proposition $p$ is considered to be \textbf{possible}, denoted $\possibility{p}$, if there exists at least one world accessible from $w_{current}$ where $p$ is true. Conversely, a proposition is considered to be \textbf{necessary}, denoted $\necessity{p}$, if $p$ is true in all worlds that are accessible from $w_{current}$. These logical rules are known to be decidable with a modal logic (without first-order operators)

\subsection{The Neuro-Symbolic Loop}
The primary innovation of our approach lies in the dynamic and interactive process by which agents update their Kripke models. This is not a static or pre-programmed process but rather a continuous loop that is actively mediated by a language model. The process unfolds in a clear, four-stage sequence:
\begin{enumerate}
    \item \textbf{Perception:} An agent observes new data from its environment, such as an anomalous sensor reading.
    \item \textbf{Hypothesis Generation (Neural):} The agent feeds this raw, unstructured data into an LM, whose task is to generate a natural-language hypothesis explaining the observation.
    \item \textbf{Logical Formulation:} This semantic hypothesis is translated into a formal logical proposition, $p_{\text{hypo}}$.
    \item \textbf{Validation \& Update (Symbolic):} The agent considers a hypothetical update to its Kripke model where $p_{\text{hypo}}$ holds true. Before committing to this change, it verifies the new hypothetical model against its set of expert knowledge axioms. 
    Only if the update does not introduce any logical contradictions will the agent transition to this new belief state, effectively pruning worlds that are no longer considered possible.
\end{enumerate}
This loop grounds the workflow where reasoning of the LM in a solid, verifiable logical structure, thereby mitigating the inherent risks of hallucination and logical inconsistency.

\subsection{Grounding Semantic Hypotheses into Logical Propositions}
 
A critical step in any neuro-symbolic architecture is the translation of a high-dimensional, semantic output from a neural model into a low-dimensional, formal symbolic representation. The paper refers to this as the "Logical Formulation" step. While this is a notoriously difficult open problem in AI, our framework addresses it through a pragmatic, domain-specific approach that prioritizes robustness and reliability over open-ended expressiveness.

The symbolic part of our system operates on a small, predefined set of atomic propositions (e.g., `cooling\_fault\_reported, `klystron\_fault\_reported,`rf\_power\_low). These propositions are explicitly defined in each agent's initial Kripke model and constitute the complete logical vocabulary for the diagnostic task. The system does not need to invent or parse novel logical terms.

The process is a constrained, multi-stage engineering solution.  First, we do not ask the LM to generate a logical formula from scratch. Instead, we use structured prompting to constrain the LM's task to one of classification. When an anomaly is detected, the LM is given a prompt that explicitly asks it to categorize the fault into one of the predefined system types. The prompt for the monitoring agents includes the following instruction:
\begin{quote}
\textit{Your response MUST be a JSON object with one key: 'suspected\_system', which should be one of the following: 'Cooling', 'Power', 'Vacuum', 'Klystron'...}
\end{quote}

This structured prompting turns the LM into a powerful semantic classifier rather than an open-ended generator with rich internal knowledge about accelerators. The LM's role is to use its world knowledge to map the semantics of an anomaly report (e.g., "high temperature") to the most likely faulty subsystem ("Cooling") from a fixed list of options.

The final translation step is a simple, deterministic mapping handled by the agent's code. The agent receives the structured JSON output from the LM and uses a hard-coded dictionary or switch-case statement to map the string output to the corresponding formal proposition. For example:
\begin{itemize}
    \item If LM returns {\{'suspected\_system': 'Cooling'\}}, the agent generates the proposition {cooling\_fault\_reported}.
    \item If LM returns {\{'suspected\_system': 'Klystron'\}}, the agent generates the proposition {klystron\_fault\_reported}.
\end{itemize}

This allows integration of the LMs with their rich knowledge for diagnosis of our complex scenario, while their outputs are guard-railed to stick to an explainable scenario. 

\section{A Multi-Agent Architecture for Diagnostics}

To effectively diagnose faults in a system as complex as a particle accelerator, a single monolithic agent is insufficient. 
We therefore employ a multi-agent system designed with a clear separation of concerns, an organizational principle that reflects the specialized nature of real-world diagnostic teams and engineering disciplines. This architecture is composed of several distinct agent types:
\paragraph{Component Monitoring Agents:} These serve as the observers, with each agent being responsible for a specific subsystem (e.g., cooling or radio-frequency power). These agents monitor a small, dedicated set of signals and utilize their local neuro-symbolic loop to form initial, localized hypotheses when anomalies are detected.
\paragraph{Hierarchical Reasoning Agent:} This agent controls the monitors and does not observe environmental signals directly. Instead, its inputs are the structured reports and belief models from the various component agents. Its primary role is to synthesize these disparate reports into a single, coherent causal theory for the entire system, and it is endowed with the most comprehensive set of expert rules.
\paragraph{Physical Knowledge Agent:} This unique agent acts as a specialized repository of factual information. It holds no beliefs about the dynamic state of the environment; its knowledge base is a static, factual representation of the system's physical topology. Its sole purpose is to answer specific queries from the Reasoning Agent to verify the physical plausibility of a given causal theory.

This collaborative process allows for a sophisticated diagnostic workflow. A component agent first detects an anomaly and forms a local hypothesis. The Reasoning Agent then correlates this hypothesis with another agent's factual report in its knowledge base. The LM within the Reasoning Agent proposes a causal link, but before this link is accepted, the Reasoning Agent formally queries the Physical Knowledge Agent to confirm that the two subsystems are physically connected, thereby grounding the entire causal theory in physical reality.

\section{The Foundational Role of Expert Knowledge}
 
A key part of our work is that for autonomous agents to be trusted in critical applications, their reasoning must be constrained by established, verifiable domain knowledge. In our framework, this is achieved by providing the central Reasoning Agent with a set of axioms formulated in modal logic. These axioms represent immutable truths about the environment that cannot, under any circumstances, be violated. These rules are not mere heuristics; they are formal logical statements that the agent's belief model, $M$, must satisfy at all times. They function as powerful logical guardrails on the LM's hypothesis generation process.

One of the most critical functions of these axioms is to constrain causal direction. Physical systems adhere to strict cause-and-effect relationships that a purely language-based model might misinterpret. Our system enforces this with an axiom such as:
$$ \necessity{} (\text{klystron\_fault\_reported} \rightarrow \text{rf\_power\_fault\_reported}) $$
This statement asserts that in all possible future worlds accessible from the current one, if a klystron fault is reported, then an RF power fault must also be reported. This enforces the physical reality that a failing power amplifier (klystron) will inevitably cause a drop in its output power. When the LM generates a hypothesis, the agent can check if this implication holds. A theory suggesting the reverse causality would be invalidated at the symbolic level, preventing the system from pursuing a physically impossible diagnostic path.

Second, the axioms enforce fundamental physical and logical constraints by declaring certain states to be impossible. This prevents the LM from conflating distinct fault classes. For example, the axiom:
$$ \necessity{} \neg (\text{cooling\_fault\_reported} \land \text{klystron\_fault\_reported}) $$
This rule states that a cooling fault and a klystron fault must not be the same event. In the context of the Kripke model, this means that for any world $w$ accessible from the current state, the valuation set $V(w)$ cannot simultaneously contain both propositions. This provides a logical boundary between different categories of failure.

Finally, expert rules can drastically improve the efficiency of the diagnostic process by pruning irrelevant hypotheses. In a complex system, many faults may be temporally correlated without being causally linked. 
An expert can define axioms to guide the agent away from these known misleading conclusions. 
For instance:
$$ \necessity{} (\text{vacuum\_fault\_reported} \rightarrow \neg \possibility{\text{rf\_fault\_is\_root\_cause}}) $$
This axiom is a powerful statement about causality. It dictates that in all accessible worlds where a vacuum fault is reported, it is not possible that an RF fault is the root cause. Semantically, within the Kripke model, this means if $V(w_{\text{current}})$ contains $\text{vacuum\_fault\_reported}$, then there can be no accessible world $w_j$ (where $(w_{\text{current}}, w_j) \in R$) for which $\text{rf\_fault\_is\_root\_cause} \in V(w_j)$. This powerfully prunes entire branches of the reasoning tree, focusing the agent's attention on physically plausible causes.

\section{Experimental Setup and Scenarios}
 
To rigorously evaluate our architecture, we developed a simulation of a simple particle accelerator sector. This environment simulates data streams from various components and, critically, models the physical coupling between them, allowing for the natural emergence of realistic cascading failures. We designed three distinct scenarios of increasing complexity to thoroughly test the system's diagnostic capabilities:
\begin{itemize}
    \item \textbf{Scenario 1: Cascading Failure.} A primary cooling system valve becomes stuck. This initially causes a pressure drop, detected by the Cooling Agent. Due to thermal loss, the connected RF cavity begins to overheat several timesteps later, an anomaly detected by the RF Agent. The expected outcome is for the system to correctly identify the initial cooling valve failure as the root cause of the subsequent temperature symptom.

    \item \textbf{Scenario 2: Direct Causal Failure.} A klystron (RF power amplifier) suffers a partial failure. This immediately causes a drop in the forward RF power, detected by another agent. This is a more straightforward test of identifying a direct causal link between two components in the same subsystem.

    \item \textbf{Scenario 3: Complex Failure with Confounding Event.} A klystron failure identical to Scenario 2 occurs. However, one timestep later, an unrelated vacuum pump in the same sector develops a fault, causing a pressure spike. This scenario tests the system's ability to perform true root cause analysis and, crucially, to avoid being misled by a temporally correlated but causally independent event.
\end{itemize}

\section{Results and Assessment}
 
The multi-agent system was executed for each scenario, demonstrating successful end-to-end diagnosis in all cases. The results validate the effectiveness of the hybrid neuro-symbolic architecture, where the LM's causal intuition is guided and verified by symbolic logic and a factual knowledge base. The analysis links neural hypothesis generation and logical validation, leading to correct and verifiable conclusions.

\subsection{Summary of Results}

\textbf{In Scenario 1 (Cascading Failure)}, the system correctly unraveled the temporally delayed causal chain. The `Cooling\_Agent` report at tick 3 caused the `AcceleratorDiagnostics` agent to update its belief state, considering worlds where a cooling fault was possible. When the `RF\_Agent` reported a temperature anomaly at tick 4 and its own LM hypothesized a cooling cause, the diagnostics agent had corroborating evidence. Its LM generated the correct causal theory, which was formalized into propositions that were validated against the expert knowledge base. 
The final, critical step was a query to the `LatticeLayoutAgent`, which confirmed the physical connection, allowing the diagnostics agent to commit to a new Kripke model state where the cooling fault was confirmed as the root cause and other possibilities were pruned.

\textbf{In Scenario 2 (Direct Causal Failure)}, the system demonstrated the power of its logical guardrails. The simultaneous faults triggered reports from both the `Klystron\_Agent` and the `RF\_Agent`. The `AcceleratorDiagnostics` agent's LM correctly proposed that the klystron was the root cause. This hypothesis was translated into a belief update that was tested against the axiom $\necessity{} (\text{klystron\_fault\_reported} \rightarrow \text{rf\_power\_fault\_reported})$. The proposed theory was logically consistent with this rule and was therefore accepted and passed to the physical verification stage. 
LM incorrectly reversed the causality; the symbolic validation would have immediately failed because the resulting belief state would have violated this necessary implication. This successful test case shows the axioms functioning as intended to ensure physically sound reasoning.

\textbf{In Scenario 3 (Complex Failure)}, the system successfully demonstrated fault isolation. It correctly diagnosed the klystron fault from tick 3 as in the previous scenario. The confounding vacuum fault introduced at tick 4 was, by design, of a small magnitude ($\Delta P < \tau_{vac}$), below the `Vacuum\_Agent`'s reporting threshold. The system correctly treated this as environmental noise, and the `AcceleratorDiagnostics` agent was never distracted by an irrelevant report. This is a crucial feature, demonstrating an ability to distinguish signals from noise. It is important to note that even if the vacuum fault had been reported, the system was protected by the axiom $\necessity{} (\text{vacuum\_fault\_reported} \rightarrow \neg \possibility{\text{rf\_fault\_is\_root\_cause}})$, which would have prevented it from ever considering the vacuum issue as a potential cause of the RF problem, showcasing a multi-layered defense against spurious correlations.

\begin{figure}[h!]
    \centering
    \includegraphics[width=\textwidth]{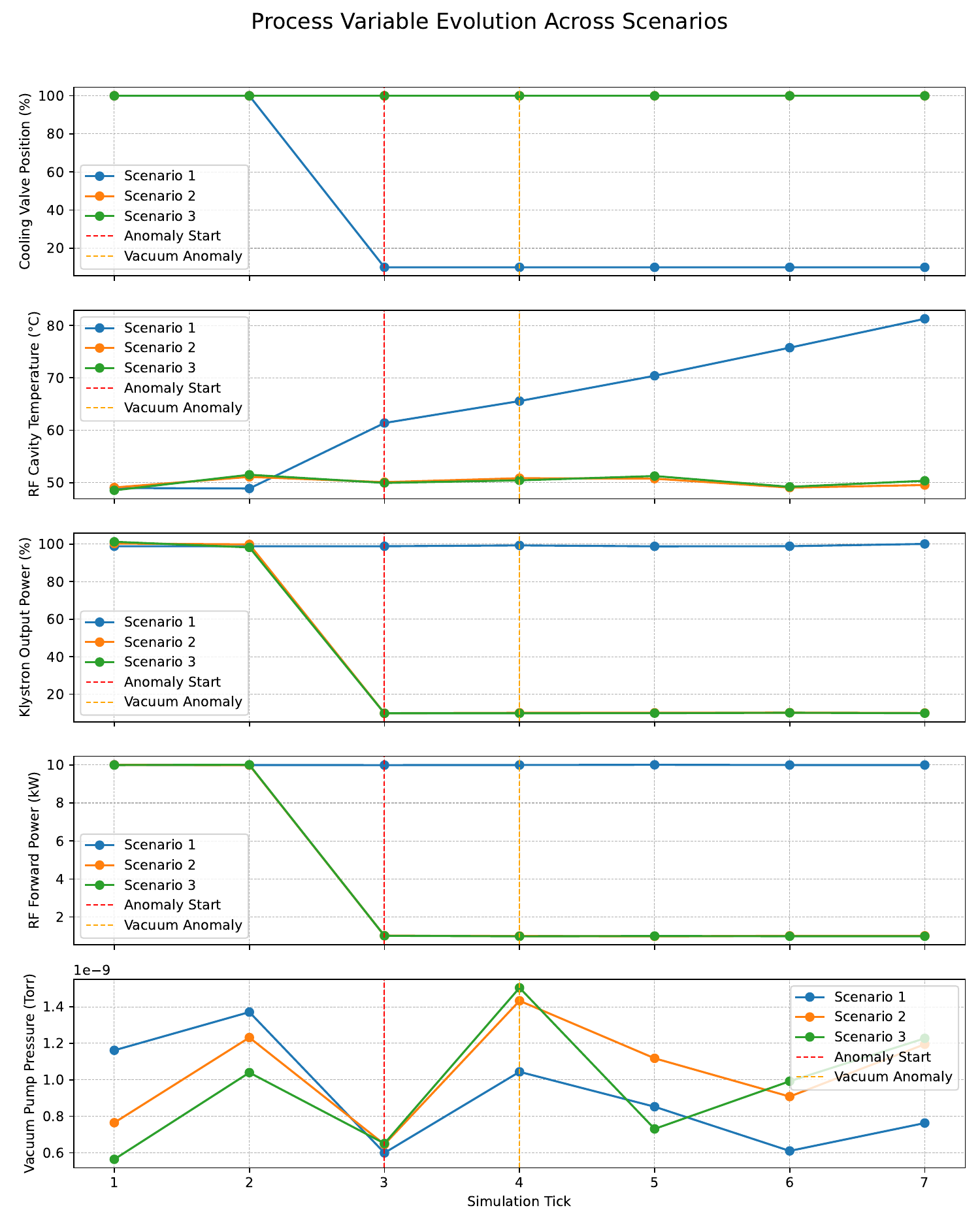}
    \caption{Time-series data for key EPICS process variables across the three diagnostic scenarios. Anomalies are introduced at Tick 3 (and Tick 4 in Scenario 3), with their effects propagating through the physically coupled system.}
    \label{fig:scenarios}
\end{figure}

\subsection{Detailed Diagnostic State Evolution}

A closer examination of the `AcceleratorDiagnostics` agent's Kripke model reveals how its beliefs evolve from uncertainty to certainty, see Fig.~\ref{fig:scenarios}. Initially, in all scenarios, the agent starts in a nominal state $w_0$, where $V(w_0) = \{\text{system\_nominal}\}$ and the accessibility relation $R$ connects $w_0$ to several possible fault worlds (e.g., cooling, klystron, vacuum), representing its initial uncertainty.

In \textbf{Scenario 1}, at Tick 3, the `Cooling\_Agent` and `RF\_Agent` reports arrive. The LM proposes a causal theory where cooling is the root cause. This is validated symbolically and physically. The agent then performs a belief update, transitioning its Kripke model. The new model, as seen at the end of the simulation, contains a new current world $w_3$, where the valuation $V(w_3)$ includes propositions like $\{\text{cooling\_insufficient}, \text{RF\_overheats}\}$. Crucially, the old nominal world $w_0$ is pruned, and the accessibility relation is updated to reflect the new, certain reality of the diagnosed fault.

In \textbf{Scenario 2}, the simultaneous reports from the `Klystron\_Agent` and `RF\_Agent` at Tick 3 lead the diagnostics agent to evaluate a transition to a world where a klystron fault is the root cause. This hypothesis is checked against the axiom $\necessity{} (\text{klystron\_fault\_reported} \rightarrow \text{rf\_power\_fault\_reported})$ and is found to be consistent. After physical verification, the agent updates its beliefs. The final Kripke model at the end of the simulation is simplified to a single world, $w_1$, where the valuation $V(w_1)$ contains propositions confirming the diagnosis, such as $\{\text{Klystron likely damaged}\}$. All other possible fault worlds have been pruned, showing a successful reduction of uncertainty to a single, confirmed diagnosis.

\subsection{Analysis}

The successful outcomes provide strong validation for the neuro-symbolic multi-agent architecture. The results highlight a deeply synergistic relationship where the neural and symbolic components enhance each other's capabilities. The primary success is the clear demonstration of a complete and correct multi-step reasoning process, integrating neural hypothesis generation, symbolic validation, and factual verification. This layered methodology ensures that any final diagnosis is not just plausible, but also logically consistent and physically possible.

The system's LM component consistently produced accurate causal theories, allowing the symbolic components to function as a robust verification layer that guarantees the integrity of the agent's reasoning. The expert knowledge axioms served as effective and respected guardrails, leading to a high degree of trust in the final diagnostic output. The successful handling of the cascading failure in Scenario 1 showcases the system's ability to correlate events across time, while the correct diagnosis in Scenario 3 demonstrates the architecture's focus and efficiency. The final Kripke models of the `AcceleratorDiagnostics` agent consistently reflected a coherent and correctly updated belief state about the environment, proving the overall effectiveness of the neuro-symbolic loop.

\section{Discussion and Future Work}
 
Our work demonstrates a promising and viable path for building more reliable and trustworthy autonomous agents for fast diagnosis at accelerator controls. 
By treating the LM not as an infallible oracle but as a powerful hypothesis generator whose outputs are subject to rigorous symbolic verification, we can effectively harness its strengths while simultaneously mitigating its known weaknesses. The use of modal logic and Kripke models provides a formal semantics for agent belief, which is a critical and necessary step towards creating agents that can transparently explain their reasoning and provide formal guarantees about their behavior. Further exploration into Dynamic Epistemic Logic could enhance the model's ability to reason about how agents' knowledge changes as a result of new information or actions~\cite{van2007dynamic}.

This framework opens up several exciting directions for future research. 
The current process of translating LM outputs into formal propositions is straightforward; developing more complex semantic parsing techniques, possibly using methodologies from the field of computational linguistics, could allow for the expression of richer and more nuanced hypotheses. 
Future work could explore advanced methods for learning these logical constraints directly from data or through guided interaction with human experts, potentially leading to systems that can formalize their own operational theories over time~\cite{runge2020discovering}.
Finally, applying this architecture to an online reinforcement learning setting, where an agent's actions have tangible consequences and its belief models are continually updated based on rich environmental feedback, represents the next interesting step toward achieving true end-to-end autonomy.

\section{Conclusion}
 
As we work toward creating more autonomous and versatile AI agents, it’s clear that they need environments that are scalable, interactive, and diverse. But scaling shouldn’t only mean bigger or more complex worlds — it must also include improving the agents’ ability to reason. For an agent to be truly reliable, it needs to build and maintain a logically consistent understanding of its environment. In this paper, we present a neuro-symbolic architecture that combines the language skills of language models (LMs) with the precision of modal logic and expert knowledge. This approach provides a framework that is both scalable and verifiable. By allowing agents to reason not just about what is true, but also about what is possible, necessary, or impossible, we take a significant step toward the long-standing goal of creating autonomous systems we can genuinely trust.

\bibliographystyle{unsrt}
\bibliography{references}

\clearpage
\appendix
 
\section{Simulation Fidelity and Simplifications}
 
The term "high-fidelity" in the context of this paper refers specifically to the simulation's ability to model the \textbf{causal and physical coupling} between accelerator components, which is the primary focus of the diagnostic task. The simulation is not intended to be a comprehensive physical model of particle beam dynamics but rather a robust testbed for the agents' reasoning capabilities.

\subsection{Fidelity Features}
\begin{itemize}
    \item \textbf{Physical Coupling}: The simulation correctly models the direction of service and dependency. For example, the {RF:forward\_power} process variable is directly dependent on the {RF:klystron\_output} variable. A failure in the klystron immediately and automatically causes a drop in the forward power, creating a realistic, direct causal link for the agents to diagnose.
    \item \textbf{Temporal Dynamics}: The simulation incorporates simplified temporal effects, such as the thermal inertia in the cooling scenario. A fault in the {COOL:valve\_position} does not cause an instantaneous temperature spike in the {RF:cavity\_temp}. Instead, the temperature rises over several simulation ticks, representing a delayed, cascading failure that is more challenging to diagnose than a simple simultaneous fault.
\end{itemize}

\subsection{Simplifications and Justifications}
To create a controlled and interpretable environment for testing logical reasoning, several simplifications were made:
\begin{itemize}
    \item \textbf{No Beam Dynamics}: The simulation does not calculate the trajectory, energy, or state of the actual particle beam. It models only the operational state of the hardware components.
    \item \textbf{Linear and Discrete Processes}: The physical relationships are simplified to linear functions, and the simulation progresses in discrete time steps ("ticks"). For instance, the temperature rise from a cooling fault is linear over time. This simplification ensures that the causal links are unambiguous for the purpose of evaluating the diagnostic logic.
    \item \textbf{Simplified Noise Models}: Sensor noise is modeled using standard uniform random distributions around a baseline value. This provides a consistent level of noise without introducing complex statistical variations that could obscure the primary fault signal.
\end{itemize}

These simplifications are justified because the experiment's goal is to validate the agents' \textbf{neuro-symbolic reasoning loop}, not to precisely replicate accelerator physics. The environment provides a clear ground truth for causality, allowing for a definitive evaluation of the diagnostic agent's conclusions.

\end{document}